\documentclass[letterpaper, 10pt, conference]{ieeeconf}
\IEEEoverridecommandlockouts
\usepackage[T1]{fontenc}\usepackage[utf8]{inputenc}
\usepackage{graphicx}\graphicspath{{figures/}}
\usepackage{amsmath,amssymb}
\usepackage{booktabs,multirow,array}
\usepackage{microtype}
\usepackage[hidelinks]{hyperref}
\usepackage{cite}
\usepackage{float}

\newcommand{\pihat}{\ensuremath{\hat{\pi}}}
\newcommand{\planner}{\ensuremath{\pi}}
\newcommand{\SD}{SparseDrive}
\newcommand{\DD}{DiffusionDrive}
\newcommand{\NCAP}{NeuroNCAP}
\newcommand{\pct}{\%}

\newcommand{\numScenes}{50}
\newcommand{\numSeeds}{3}

\newcommand{\failPlanning}{95.9\%}        

\newcommand{\sdStockNcap}{4.009}\newcommand{\sdStockCR}{28.0}
\newcommand{\sdStockStat}{4.935}\newcommand{\sdStockFront}{3.692}\newcommand{\sdStockSide}{3.399}
\newcommand{\sdStockChalCR}{1.00}\newcommand{\sdStockChalL}{1.66}
\newcommand{\sdStockNuscCR}{0.094}\newcommand{\sdStockNuscL}{0.609}
\newcommand{\ddStockNcap}{3.624}\newcommand{\ddStockCR}{37.5}
\newcommand{\ddStockStat}{4.806}\newcommand{\ddStockFront}{3.220}\newcommand{\ddStockSide}{2.847}
\newcommand{\ddStockChalCR}{1.67}\newcommand{\ddStockChalL}{2.37}
\newcommand{\ddStockNuscCR}{0.079}\newcommand{\ddStockNuscL}{0.574}

\newcommand{\sdOursNcap}{4.639}\newcommand{\sdOursNcapStd}{0.04}
\newcommand{\sdOursCR}{12.2}\newcommand{\sdOursCRStd}{0.8}
\newcommand{\sdOursStat}{5.000}\newcommand{\sdOursFront}{4.010}\newcommand{\sdOursSide}{4.907}
\newcommand{\sdOursChalCR}{0.86}\newcommand{\sdOursChalCRStd}{0.03}\newcommand{\sdOursChalL}{1.68}\newcommand{\sdOursChalLStd}{0.01}
\newcommand{\sdOursNuscCR}{0.088}\newcommand{\sdOursNuscCRStd}{0.020}\newcommand{\sdOursNuscL}{0.668}\newcommand{\sdOursNuscLStd}{0.070}
\newcommand{\ddOursNcap}{4.394}\newcommand{\ddOursNcapStd}{0.19}
\newcommand{\ddOursCR}{19.2}\newcommand{\ddOursCRStd}{4.6}
\newcommand{\ddOursStat}{4.967}\newcommand{\ddOursFront}{3.861}\newcommand{\ddOursSide}{4.354}  
\newcommand{\ddOursChalCR}{1.02}\newcommand{\ddOursChalCRStd}{0.17}\newcommand{\ddOursChalL}{1.98}\newcommand{\ddOursChalLStd}{0.09}
\newcommand{\ddOursNuscCR}{0.071}\newcommand{\ddOursNuscCRStd}{0.008}\newcommand{\ddOursNuscL}{0.580}\newcommand{\ddOursNuscLStd}{0.005}
\newcommand{\sdChalNuscCR}{0.135}\newcommand{\sdChalNuscCRStd}{0.029}\newcommand{\sdChalNuscL}{0.689}\newcommand{\sdChalNuscLStd}{0.070}

\newcommand{\sdChalNcap}{4.276}\newcommand{\sdChalNcapStd}{0.04}\newcommand{\sdChalCR}{19.1}\newcommand{\sdChalCRStd}{0.9}
\newcommand{\sdChalStat}{4.839}\newcommand{\sdChalFront}{4.282}\newcommand{\sdChalSide}{3.708}
\newcommand{\sdChalChalCR}{0.87}\newcommand{\sdChalChalCRStd}{0.08}\newcommand{\sdChalChalL}{1.55}\newcommand{\sdChalChalLStd}{0.03}
\newcommand{\ddChalNcap}{3.581}\newcommand{\ddChalNcapStd}{0.21}\newcommand{\ddChalCR}{38.8}\newcommand{\ddChalCRStd}{4.5}
\newcommand{\ddChalStat}{4.764}\newcommand{\ddChalFront}{2.998}\newcommand{\ddChalSide}{2.983}
\newcommand{\ddChalChalCR}{0.63}\newcommand{\ddChalChalCRStd}{0.03}\newcommand{\ddChalChalL}{1.22}\newcommand{\ddChalChalLStd}{0.13}
\newcommand{\ddChalNuscCR}{0.080}\newcommand{\ddChalNuscCRStd}{0.030}\newcommand{\ddChalNuscL}{0.580}\newcommand{\ddChalNuscLStd}{0.007}
\newcommand{\sdRuleNcap}{4.373}\newcommand{\sdRuleNcapStd}{0.13}\newcommand{\sdRuleCR}{17.9}\newcommand{\sdRuleCRStd}{1.0}
\newcommand{\sdRuleChalCR}{0.91}\newcommand{\sdRuleChalL}{1.60}\newcommand{\sdRuleNuscCR}{0.116}\newcommand{\sdRuleNuscL}{0.652}
\newcommand{\ddRuleNcap}{3.759}\newcommand{\ddRuleNcapStd}{0.06}\newcommand{\ddRuleCR}{34.3}\newcommand{\ddRuleCRStd}{1.5}
\newcommand{\sdRuleChalCRStd}{0.05}\newcommand{\sdRuleChalLStd}{0.05}\newcommand{\sdRuleNuscCRStd}{0.014}\newcommand{\sdRuleNuscLStd}{0.003}
\newcommand{\ddRuleChalCRStd}{0.05}\newcommand{\ddRuleChalLStd}{0.03}\newcommand{\ddRuleNuscCRStd}{0.016}\newcommand{\ddRuleNuscLStd}{0.009}
\newcommand{\sdRuleStat}{4.843}\newcommand{\sdRuleFront}{4.134}\newcommand{\sdRuleSide}{4.141}

\newcommand{\ddRuleStat}{4.876}\newcommand{\ddRuleFront}{3.296}\newcommand{\ddRuleSide}{3.104}

\newcommand{\ddRuleChalCR}{0.78}\newcommand{\ddRuleChalL}{1.38}\newcommand{\ddRuleNuscCR}{0.095}\newcommand{\ddRuleNuscL}{0.580}

\newcommand{\sdRecNcap}{4.144}\newcommand{\sdRecCR}{23.8}\newcommand{\sdRecStat}{4.999}\newcommand{\sdRecFront}{3.815}\newcommand{\sdRecSide}{3.619}
\newcommand{\sdRecChalCR}{0.94}\newcommand{\sdRecChalL}{1.66}\newcommand{\sdRecNuscCR}{0.114}\newcommand{\sdRecNuscL}{0.636}
\newcommand{\sdRecNcapStd}{0.07}\newcommand{\sdRecCRStd}{2.5}
\newcommand{\sdRecChalCRStd}{0.05}\newcommand{\sdRecChalLStd}{0.06}\newcommand{\sdRecNuscCRStd}{0.006}\newcommand{\sdRecNuscLStd}{0.013}
\newcommand{\ddRecNcap}{3.700}\newcommand{\ddRecCR}{34.7}
\newcommand{\ddRecStat}{4.702}\newcommand{\ddRecFront}{3.277}\newcommand{\ddRecSide}{3.119}
\newcommand{\ddRecChalCR}{0.86}\newcommand{\ddRecChalL}{1.53}
\newcommand{\ddRecNuscCR}{0.090}\newcommand{\ddRecNuscL}{0.589}
\newcommand{\ddRecNcapStd}{0.16}\newcommand{\ddRecCRStd}{3.7}
\newcommand{\ddRecChalCRStd}{0.07}\newcommand{\ddRecChalLStd}{0.09}
\newcommand{\ddRecNuscCRStd}{0.006}\newcommand{\ddRecNuscLStd}{0.009}

\newcommand{\ddBrakeNcap}{3.833}\newcommand{\ddBrakeCR}{33.1}
\newcommand{\ddBrakeChalCR}{1.32}\newcommand{\ddBrakeChalL}{1.95}

\newcommand{\fsOursNcap}{3.939}\newcommand{\fsOursCR}{30.8}
\newcommand{\fsOursChalCR}{1.00}\newcommand{\fsOursChalL}{1.75}

\newcommand{\sdCrossNcap}{4.460}\newcommand{\sdCrossCR}{15.7}
\newcommand{\ddCrossNcap}{3.837}\newcommand{\ddCrossCR}{30.8}

\newcommand{\premiumSD}{0.18}\newcommand{\premiumDD}{0.56}

\newcommand{\fidEligClean}{5{,}119}  
\newcommand{\sdBarClean}{0.609}\newcommand{\ddBarClean}{0.575}
\newcommand{\sdPihatClean}{0.620}\newcommand{\ddPihatClean}{0.643}
\newcommand{\sdPihatCleanX}{1.02}\newcommand{\ddPihatCleanX}{1.12}

\newcommand{\sdPrecClean}{0.603}\newcommand{\ddPrecClean}{0.673}  

\newcommand{\sdBarCR}{1.60}\newcommand{\ddBarCR}{1.33}
\newcommand{\sdPihatCR}{1.52}\newcommand{\ddPihatCR}{1.07}
\newcommand{\sdRecallClean}{0.573}\newcommand{\ddRecallClean}{0.544}  
\newcommand{\sdLift}{38$\times$}\newcommand{\ddLift}{51$\times$}

\newcommand{\std}[1]{{\scriptsize$\pm$#1}}

\usepackage{xcolor}

\title{\LARGE \bf Model-Aware Generation of Safety-Critical Scenes for Robust End-to-End Driving via Distilled Planner Surrogates}
\author{
Khaled M. Seyam$^{1}$, Julian Wiederer$^{2}$, Samar Fathalah$^{1}$, Markus Braun$^{2}$, Bin Yang$^{1}$\\
$^{1}$Institute of Signal Processing and System Theory, University of Stuttgart, Germany\\
$^{2}$Mercedes-Benz AG, Stuttgart, Germany\\
}
\begin{document}
\maketitle
\thispagestyle{empty}\pagestyle{empty}

\begin{abstract}
End-to-end driving models perform well on standard benchmarks but collide frequently in safety-critical scenarios. We present a failure-mode analysis on NeuroNCAP which shows that \failPlanning{} of DiffusionDrive's collisions are planning failures despite reliable detection, pinpointing the bottleneck in planning rather than perception. We attribute this to a fundamental data imbalance, as safety-critical interactions are rare in real driving logs, so planners are hardly ever trained on them. We close this gap with a novel model-aware framework that generates safety-critical training scenes against a specific planner. Each planner is attacked at the bird's-eye-view (BEV) level through a differentiable surrogate distilled from its own planning head. This keeps the expensive video renderer out of the attack loop. 
We extract \numScenes{} scenes per planner, each pairing an attack with a safe escape trajectory that serves as the fine-tuning label.
A fully automated quality-diversity search generates both the attacks and the escapes, so that their variety is not specified by hand.
Only these scenes are rendered to fine-tune the planner, which reduces the computational cost significantly. 
Fine-tuning on these scenes roughly halves the closed-loop collision rate of two planners on NeuroNCAP (\SD{} \sdStockCR\pct$\to$\sdOursCR\pct, \DD{} \ddStockCR\pct$\to$\ddOursCR\pct) while largely preserving normal driving. A cross-planner transfer experiment shows that the gain comes from targeting the specific planner. Each planner improves most on the scenes generated against itself, and \DD{} fine-tuned on \SD{}'s scenes retains only 6.7 of its 18.3-point collision-rate reduction.
\end{abstract}

\section{Introduction}
End-to-end (E2E) autonomous driving has shown strong performance on standard benchmarks, yet the models are limited by the data they are trained on.
Real-world driving datasets are heavily skewed toward normal, uneventful driving, with safety-critical interactions accounting for less than 0.03\% of real-world driving time~\cite{wode2e}. Such situations are inherently rare in naturalistic driving and cannot be deliberately staged at scale for safety, cost, and legal reasons.
As soon as these models encounter scenarios outside their training distribution, performance degrades sharply.
On Challenger, which generates physically plausible videos for adversarial scenarios such as cut-ins, sudden lane changes, and blind-spot intrusions, the collision rate of state-of-the-art E2E models increases by up to 26$\times$ compared to standard nuScenes evaluation~\cite{challenger}.
While Challenger stress-tests models through video-based adversarial replay, NeuroNCAP evaluates in closed-loop, where the ego executes its own plan and failures hidden in open loop appear, on harder scenarios including oncoming vehicles, stationary vehicles, and side collisions~\cite{neuroncap}.
There \SD{} results in a collision rate of \sdStockCR\pct{} and \DD{} in \ddStockCR\pct.

\begin{figure}[!t]
  \centering
  \includegraphics[width=\columnwidth]{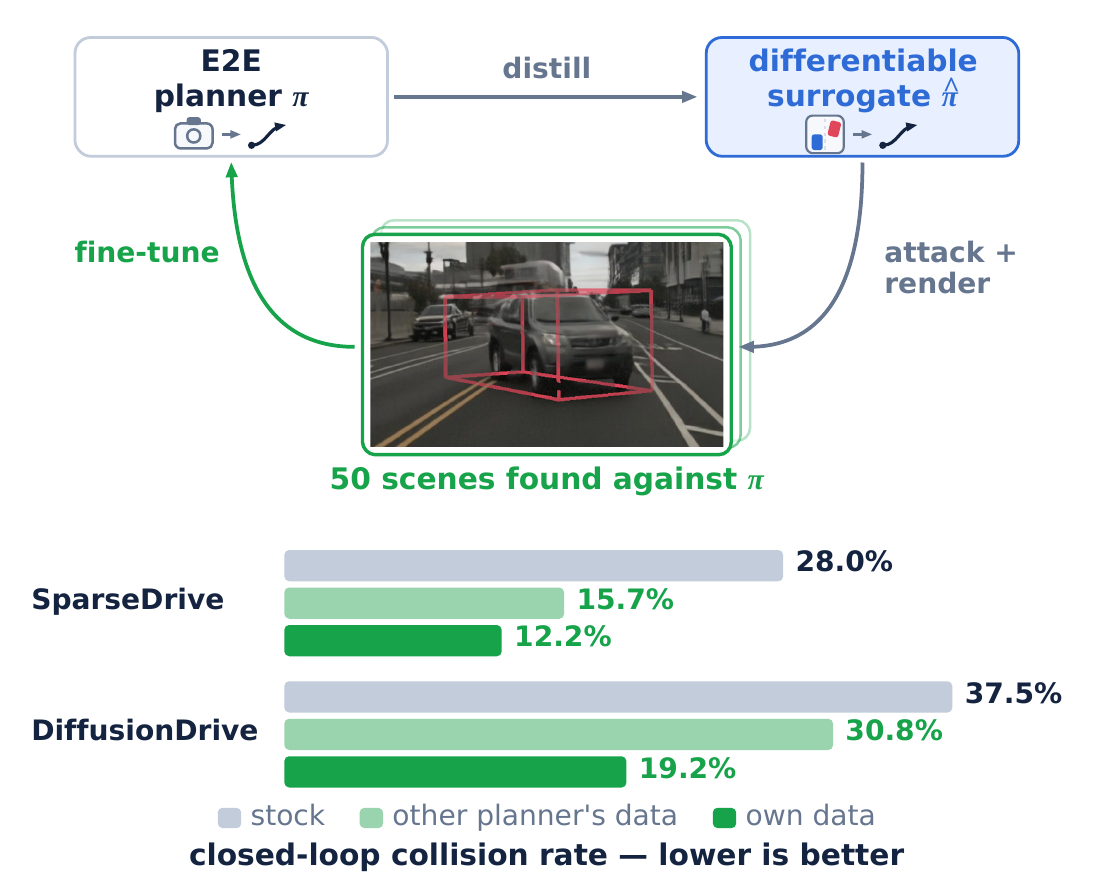}
  \vspace{-10pt}
  \caption{\textbf{Model-aware adversarial fine-tuning.} We distill an end-to-end planner into a differentiable surrogate, optimize adversarial scenarios against it, and render them to fine-tune that planner. The renderer never enters the optimization loop. Bottom: closed-loop collision rate on NeuroNCAP. Each planner improves most on the scenes generated against itself.}
  \label{fig:teaser}
\end{figure}

Since collecting safety-critical driving data at scale remains impractical, several directions have been explored to synthetically close this gap.
Video generation models such as MagicDriveV2~\cite{magicdriveV2} can synthesize photorealistic multi-view driving videos.
Yet they are evaluated almost exclusively on perceptual quality metrics like FID and FVD, and their videos have so far served perception tasks such as 3D detection and tracking~\cite{cosmos, drivedreamer2} rather than planning.
Methods that do generate training data for the planner either derive it from validation failures without closed-loop evaluation~\cite{delphi} or simulate hundreds of thousands of scenes reconstructed from logged drives~\cite{simscale}.
In both cases the scenarios come from existing data rather than being designed as interactions.
Adversarial scenario generation designs such interactions deliberately, but either at the state level without photorealistic rendering~\cite{strive,safesim,king} or, as in Challenger~\cite{challenger}, against the logged ego trajectory rather than against a planner.
No existing method generates photorealistic adversarial scenes optimized against a specific vision-based planner.

To locate the bottleneck behind these E2E failures, we decompose every closed-loop collision on NeuroNCAP into a perception or a planning failure by checking whether the planner detected and tracked the adversary.
In \failPlanning{} of DiffusionDrive's collisions, the threat is detected, so the model sees the danger but fails to react.
Targeting this weakness with generated scenes requires the gradient of a collision objective with respect to the adversary's behavior.
For a vision-based planner, whose input is rendered video, this gradient would have to pass through the video renderer, which is computationally infeasible.
We therefore attack a surrogate instead.
We distill the target planner $\planner$ into a differentiable surrogate planner $\pihat$, which consists of the planner's own planning head, copied and frozen, behind a small trained adapter that takes ground-truth bird's-eye-view (BEV) state instead of images.
The failure analysis justifies this construction, since perception is reliable.
Gradients through $\pihat$ are cheap and point at the planner's specific failure modes rather than at a generic driving prior.
Every generated scene is therefore optimized against the planner to be fine-tuned.

Those gradients drive the attack search, steering an adversarial vehicle in BEV to maximize collision risk for the planner.
A quality-diversity search spreads the attacks over different behaviors, so the types of attack are found by the search rather than specified by hand.
A geometric solver then computes a safe escape trajectory for every scene as its fine-tuning label and rejects scenes for which no escape exists.
The search costs around 30 GPU-hours per planner, and only the \numScenes{} surviving scenes are later rendered photorealistically.
We evaluate closed-loop on NeuroNCAP, open-loop on Challenger's adversarial replay, and on clean nuScenes to confirm that fine-tuning does not degrade normal driving.
These \numScenes{} scenes roughly halve each planner's closed-loop collision rate on NeuroNCAP (\SD{} \sdStockCR\pct{} to \sdOursCR\pct, \DD{} \ddStockCR\pct{} to \ddOursCR\pct; Fig.~\ref{fig:teaser}) while normal driving is largely preserved.

In short, our contributions are:
\begin{itemize}
    \item A failure-mode analysis on NeuroNCAP showing that \failPlanning{} of DiffusionDrive's closed-loop collisions are planning failures despite reliable detection, which motivates fine-tuning the planning head only.
    \item A novel differentiable surrogate of a vision-based E2E planner, built from the planner's own frozen planning head behind a small adapter over ground-truth BEV state and validated against the real planner.
    \item To our knowledge, the first model-aware pipeline for generating photorealistic safety-critical scenes for vision-based E2E planners. It optimizes attacks through the surrogate with a quality-diversity search, labels each scene with a geometrically solved escape, and rejects scenes for which no escape exists. With \numScenes{} scenes per planner, it roughly halves the closed-loop collision rate of two planners.
    \item A cross-planner transfer experiment showing that each planner improves most on the scenes generated against itself, and that this benefit is larger for \DD{} than for \SD{}. The gain therefore comes from targeting the specific planner rather than from adversarial data in general.
\end{itemize}
We will release the code, the distilled surrogates, and the generated adversarial scenes for both planners.
\section{Related Work}

\noindent\textbf{End-to-End Autonomous Driving.}
End-to-end driving models map raw sensor observations directly to planning actions, bypassing modular pipelines~\cite{chen2024e2e}.
Recent planners include imitation-based architectures~\cite{uniad, vad}, diffusion-based planners that model multimodal trajectory distributions~\cite{diffusiondrive}, and trajectory scoring approaches~\cite{sparsedrive}.
Strong scores on standard benchmarks~\cite{navsim} say little about their behavior in rare interactions, and since such interactions barely appear in real logs~\cite{wode2e}, the training data cannot correct this on its own.

\noindent\textbf{Adversarial Scenario Generation.}
Several methods optimize traffic scenarios against a planner.
STRIVE~\cite{strive} perturbs the latents of a learned traffic prior until a rule-based planner crashes, SAFE-SIM~\cite{safesim} guides trajectory diffusion with collision costs, CTG++~\cite{ctgpp} steers scene-level diffusion with loss functions derived from language queries, CAT~\cite{cat} runs closed-loop adversarial training in state-level simulation, AdvSim~\cite{advsim} perturbs actor trajectories and re-simulates LiDAR sweeps, and KING~\cite{king} obtains attack gradients through a kinematic proxy of the simulator dynamics in CARLA.
More recent methods steer language models or learned motion priors toward critical interactions~\cite{llmattacker, advbmt}.
All of these act at the state or LiDAR level against rule-based, privileged, or simulator-native planners, and none renders its attacks photorealistically for a vision-based planner.
Our adversary is instead optimized against the deployed vision-based planner itself, through a distilled surrogate of its own planning head.
The resulting scenes reach that planner as photorealistic video.

\noindent\textbf{Generative Driving Data and Photorealistic Evaluation.}
Generative video models expand driving data distributions, from structured-controllability work~\cite{wovogen, drivedreamer2, svsgan} to MagicDriveV2~\cite{magicdriveV2}, which synthesizes high-resolution multi-camera video with fine-grained control over scene layout via 3D bounding boxes and trajectories, making it a suitable backbone for targeted scenario synthesis.
Generated video has mostly served perception tasks, with quality measured by FID and FVD rather than downstream planning.
Delphi~\cite{delphi} guides generation with validation-set failure patterns, which risks benchmark-specific overfitting and is never evaluated closed-loop. SimScale~\cite{simscale} improves planner robustness through sim-real co-training on 147K recovery-based simulated scenes reconstructed from logged drives rather than designed as interactions.
On the evaluation side, NeuroNCAP~\cite{neuroncap} tests vision-based planners closed-loop in neurally rendered collision scenarios and Challenger~\cite{challenger} renders adversarial video for open-loop stress testing, and we adopt both as benchmarks.

\noindent\textbf{Surrogates and Attack Transfer.}
\begin{figure*}[!t]
  \centering
  \includegraphics[width=\textwidth]{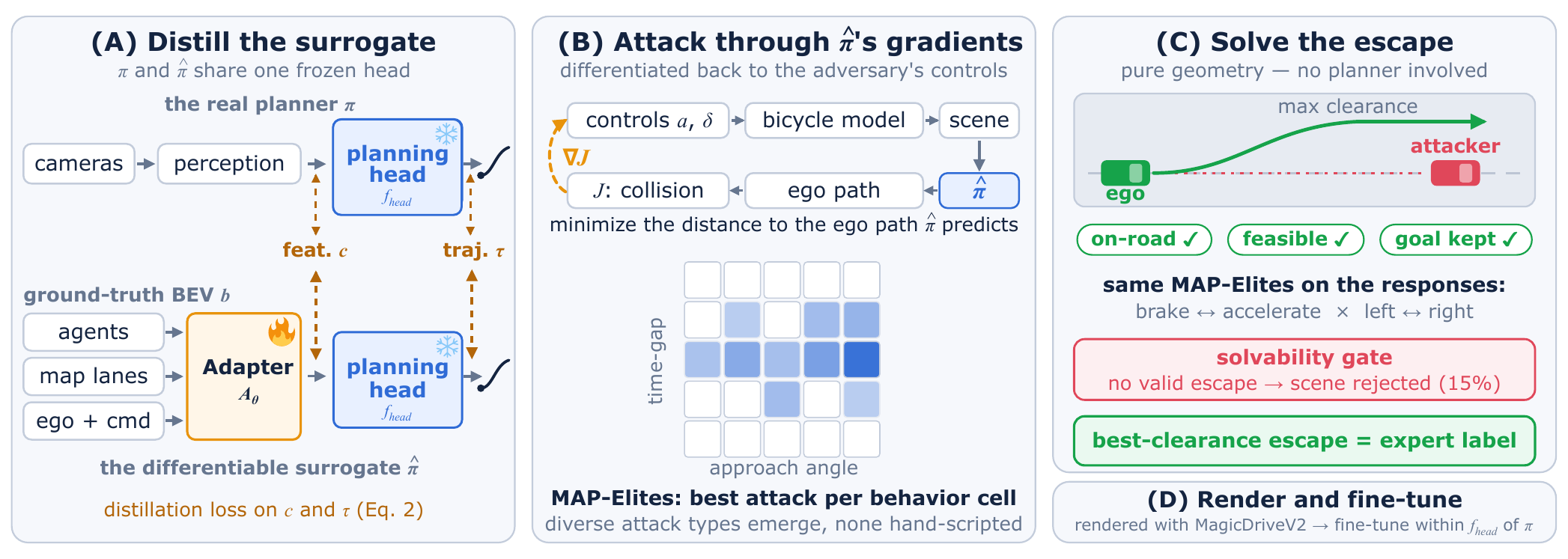}
  \vspace{-10pt}
\caption{\textbf{Method.} (A)~Only the adapter $A_\theta$ trains; the planner's own planning head $f_\mathrm{head}$ is copied and frozen, so gradients through $\pihat$ point at the planner's failure modes. (B)~Adversary controls $(a,\delta)$ are optimized against the collision objective $J$ through the surrogate, and a quality-diversity grid keeps the attack closest to the ego per cell, so attack types are found rather than specified. (C)~The escape is solved by geometry with the attacker held fixed, and scenes with no valid escape are rejected. (D)~Surviving scenes are rendered and only parameters inside the planning head of $\planner$ are updated.}
  \label{fig:pipeline}
\end{figure*}
Attacking a substitute model is standard in adversarial machine learning, where transferability across models is well documented but partial~\cite{demontis}, and STRIVE's rule-based proxy planner is the scenario-generation analogue.
Our surrogate is not a generic proxy, as it reuses the target's own frozen planning head. So its gradients are specific to that planner, and our transfer experiment measures the effect of that specificity.
To our knowledge, this is the first work to distill a vision-based end-to-end planner into a differentiable surrogate and generate its own training attacks through it.

\section{Method}\label{sec:method}
A vision-based end-to-end planner $\planner$ maps surround-view images through a perception module to context features $c$, which a planning head $f_{\mathrm{head}}$ decodes into the ego trajectory $\tau$.
Given a trained $\planner$ and real driving logs, we seek a small set of safety-critical training scenes.
Each scene is a photorealistic video of an adversarial vehicle paired with an expert label for the ego's escape, and fine-tuning $\planner$ on these scenes should reduce its closed-loop collision rate without degrading normal driving.
The generation of the scenes is model-aware in the sense that every attack is optimized against $\planner$'s own planning head. Thus two different planners receive different safety-critical scenes.
The pipeline runs in four stages (Fig.~\ref{fig:pipeline}): (A)~distill $\planner$ into a differentiable surrogate that plans from ground-truth BEV state, (B)~optimize adversarial trajectories through the surrogate under a quality-diversity search, (C)~solve a safe escape for every attack and reject scenes that admit none, and (D)~render the surviving scenes photorealistically and fine-tune the real planner on them.

\subsection{Distilling a Differentiable Surrogate}\label{sec:surrogate}
Optimizing an adversary against the planner $\planner$ requires the gradient of a collision objective with respect to the adversary's motion, taken through the planner.
Since $\planner$ consumes images, that puts the video renderer inside the optimization loop.
Rendering a single scene, a six-camera clip of 65 frames at 12 Hz (5.4 s) and 848×1600 pixels per camera, already costs about 1.4 GPU-hours, one attack needs on the order of sixty renders, and each optimization step would additionally have to differentiate through the video renderer.
Attacking the real planner directly is therefore computationally out of reach. We build a surrogate that maps the BEV state instead of images.

The surrogate keeps the planner's decision logic and replaces only its input path,
\begin{equation}
\pihat(b) = f_{\mathrm{head}}\!\left(A_\theta(b)\right),
\label{eq:pihat}
\end{equation}
where $f_{\mathrm{head}}$ is the planner's own planning head with weights copied and frozen, and $A_\theta$ is a differentiable adapter that maps the ground-truth BEV scene state $b$ into the space of context features $c$.
The state $b$ collects agent boxes with class and motion history, map lanes, the ego state and the navigation command.
$A_\theta$ is deliberately small, 0.83\,M parameters against 86.2 M for \SD{} and 92.1 M for \DD{}.
It uses one encoder per input group (agents, map lanes, ego state), and each encoder produces one part of the features $c$.

Training data comes from the planner itself.
We run $\planner$ on nuScenes and record, per sample, the context features $c$ its perception produces and the trajectory $\tau$ its head outputs, while $b$ is assembled from the same samples' ground-truth annotations.
Only $\theta$ trains using the loss
\begin{equation}
\mathcal{L}(\theta) = \lambda_{\mathrm{f}}\,\bigl\lVert A_\theta(b) - c \bigr\rVert_2^2
\;+\; \lambda_{\mathrm{t}}\,\bigl\lVert \pihat(b) - \tau \bigr\rVert_2^2 ,
\label{eq:distill}
\end{equation}
so the adapter is matched to the head twice, once in the feature space the head reads and once in the trajectory the head produces from those features.

For DiffusionDrive, whose head scores several trajectory modes and executes the best, the mode selection is softened so the forward path stays differentiable.
We validate the construction before any use.
First, we confirm that the extracted head reproduces the real planner's trajectory when given the recorded features.
Second, the gradient of a distance-to-ego-path objective with respect to the adversary's controls, computed through the surrogate, is finite and has a similar magnitude across scenes for both planners.
The construction rests on the assumption that ground-truth BEV approximates what the planner actually perceives.
Our failure analysis supports this hypothesis, as \failPlanning{} of \DD{}'s (95.8\% of \SD{}'s) closed-loop collisions occur with perception working (Sec.~\ref{sec:failure}).
\subsection{Model-Aware Attack Generation}\label{sec:attack}
An attack is a sequence of controls for one adversarial vehicle, an acceleration $a$ and a steering angle $\delta$ at each of the $T$ timesteps of the scene.
A kinematic bicycle model unrolls the controls into a trajectory, and replacing the adversary's original trajectory in $b$ with it gives the attacked BEV state $b(a,\delta)$.
We minimize a collision objective $J(a,\delta)$ by gradient descent on the controls through the bicycle model and the surrogate.
$J(a,\delta)$ consists of several terms.
The main term draws the adversary toward the ego path that $\pihat$ predicts for $b(a,\delta)$.
A timing term forces the attacker and the ego to reach the meeting point at the same time.
Further terms keep the attack on the drivable area, clear of other agents, and within realistic steering, acceleration and jerk.
Optimizing controls rather than positions keeps every attack physically drivable.
The controls for the full horizon are optimized jointly in a single gradient run.
During the run, $\pihat$ is evaluated at several frames along the rolled-out trajectory, so the ego's predicted reaction stays consistent with the attacker's committed path.
Because $\pihat$ re-plans as the adversary moves, the attack is dynamically coupled to what this planner would actually do rather than to a static script.

A single optimized attack, however, tends to rediscover the same failure.
To cover a planner's failure modes without hand-picking scenario types, we run the optimization inside MAP-Elites~\cite{mapelites}, a quality-diversity search.
MAP-Elites divides a space of behavior descriptors into a grid and keeps, in each grid cell, the attack that comes physically closest to the ego. 
We use two descriptors, contact time-gap and approach angle, on a $5\times5$ grid.
The grid forces the attacks to spread over the descriptor space, while the type of attack in each cell, such as a cut-in or a head-on approach, is not specified and emerges from the optimization.
We run one grid per base scene and keep, among its filled cells, the attack with the smallest ego--attacker distance, so each base scene contributes one attack.
The resulting distribution over attack types differs between the two planners (Sec.~\ref{sec:diversity}).
Generating a dataset of \numScenes{} scenes costs around 30 GPU-hours of search per planner, plus 70 for rendering the survivors (Sec.~\ref{sec:ft}). Even a gradient-free search with the real planner in the loop would need one render per candidate, roughly 4{,}300 GPU-hours. Our surrogate planner, together with the automated search of diverse attacks before any scene is rendered, therefore reduces the search cost substantially.

\subsection{Solving Safe Escapes}\label{sec:escape}
The attack defines the situation; the escape defines the correct response to it.
Holding the attacker's trajectory fixed, we solve for an escape, the ego trajectory that maximizes clearance to the adversary while staying on-road, avoiding all other agents, remaining kinematically feasible, and keeping the route goal.
No planner participates in this stage.
We run the same MAP-Elites search over the escapes, with a longitudinal descriptor (brake to accelerate) and a lateral one (left to right) on a $5\times5$ grid.
It returns several qualitatively distinct escapes per scene, and the escape with the largest clearance serves as the label.
Any scene where no trajectory clears the adversary by 4\,m after a 1 s (12-frame) reaction delay is rejected as unavoidable (15\% of attacks). Thus, the planner is never fine-tuned on adversarial situations it could not have survived.

\subsection{Rendering and Fine-Tuning}\label{sec:ft}
To prepare each surviving scene for rendering, all scene tensors (BEV maps, 3D bounding boxes, camera visibility masks and relative pose embeddings) are recomputed for spatial consistency with the attacker's trajectory and the ego's escape.
Each scene is then rendered with MagicDriveV2~\cite{magicdriveV2} into a six-camera photorealistic clip of 65 frames at $848\times1600$ per camera.
The base scenes come from the nuScenes train split, so no evaluation scene is used. The renderer itself is trained only on nuScenes.
Rendering takes about 1.4 GPU-hours per scene, around 70 GPU-hours for the \numScenes{} scenes of one planner.
We fine-tune the E2E model on the \numScenes{} rendered scenes mixed with clean nuScenes, so normal driving is not forgotten.
Perception stays frozen and only parameters inside the planning head $f_\mathrm{head}$ are updated.
The surrogate is used only to find the \numScenes{} scenes and is not deployed.

\section{Experiments}\label{sec:experiments}
\subsection{Setup}\label{sec:exp-setup}
We evaluate two vision-based E2E planners from their released checkpoints, \SD{}~\cite{sparsedrive} and \DD{}~\cite{diffusiondrive}. Both planners are evaluated as released, with their own inference pipelines unchanged.
NeuroNCAP~\cite{neuroncap} provides closed-loop evaluation in photorealistic collision scenarios of three categories (stationary, frontal, side), reporting a score ($\uparrow$, macro-averaged over categories) and a collision rate ($\downarrow$).
Challenger~\cite{challenger} provides open-loop adversarial replay.
Its adversaries are optimized to come close to the ego without colliding, and scenes that the logged ego trajectory cannot traverse safely are discarded.
The logged trajectory serves as ground truth, so the ground-truth ego passes close to the adversary without reacting to it.
Challenger renders with the same generative model as our pipeline, MagicDriveV2, while \NCAP{} uses neural reconstruction of the original scenes.
Clean nuScenes val~\cite{nuscenes} checks that normal driving is preserved.
Fine-tuning runs use three seeds and we report mean and standard deviation.

We compare against three data-generation baselines.
The first is recovery-based augmentation inspired by SimScale~\cite{simscale}, where ego trajectories are perturbed within the nuScenes vocabulary and recovery trajectories are retrieved from a large human-driven trajectory bank via pose-matching.
The second is rule-based, which places scripted stationary and oncoming adversaries on the ego's route, with escape labels produced by a CEM-Diffusion expert. It is safety-critical by design but blind to the planner.
The third is Challenger-style, which runs Challenger's own generation on \numScenes{} nuScenes train scenes. Its adversaries are optimized to pass close to the ego without colliding, and the logged ego trajectory serves as label, so the data is adversarial but neither aimed at a planner nor paired with an evasive label.
All three baselines use MagicDriveV2~\cite{magicdriveV2} for video synthesis and receive the same budget of \numScenes{} scenes and the same fine-tuning recipe as ours.
So the comparison isolates the scenario design strategy rather than the renderer.

The adapter trains for 20 epochs on the nuScenes train split with AdamW at learning rate $1\times10^{-4}$, weight decay 0.01 and batch size 8, weighting the two terms of Eq.~\ref{eq:distill} at $\lambda_{\mathrm{f}}=1$ and $\lambda_{\mathrm{t}}=0.1$.
Each attack search fills a $5\times5$ MAP-Elites grid over contact time-gap and approach angle with a budget of 60 optimizations per scene, each 40 Adam steps, stopping early after 15 optimizations without improvement.
The escape solver uses the same grid size and budget over its response descriptors.

The \numScenes{} generated scenes are repeated eight times in the training mix, so that they make up approximately 13.6\% of the training samples.
For both models, the perception stack is fully frozen and only layers inside the planning head $f_\mathrm{head}$ of Eq.~\ref{eq:pihat} are trained. For \DD{}, the motion-interact layers train at a base learning rate of $1\times10^{-5}$ and the diffusion branch at half of that rate.
For \SD{}, the planning output branches train at a rate of $1\times10^{-4}$ with batch size 8.
Both models train for 30 epochs with cosine annealing.
The detection, map and motion losses are disabled, so only the planning loss is active.
\begin{table}[!t]
\centering
\caption{Failure decomposition on NeuroNCAP (stock \DD{}). An agent counts as detected at recall@5--35\,m~$\geq0.5$ on the model's own detections. Rates weight every episode equally, unlike \NCAP{}'s family-weighted collision rate.}
\label{tab:failure}
\setlength{\tabcolsep}{4pt}
\footnotesize
\begin{tabular}{lccccc}
\toprule
Scenario & Det.+Cr. & Det.+No Cr. & Miss+Cr. & Miss+No Cr. & Recall \\
\midrule
Stationary &  6.8\% & 82.9\% & 0.0\% & 10.2\% & 0.822 \\
Frontal    & 52.5\% & 41.7\% & 4.5\% &  1.2\% & 0.861 \\
Side       & 50.0\% & 50.0\% & 0.0\% &  0.0\% & 0.915 \\
\midrule
Overall    & 27.8\% & 65.3\% & 1.2\% &  5.7\% & 0.866 \\
\bottomrule
\end{tabular}
\end{table}

\subsection{Failure Analysis of E2E Models}\label{sec:failure}
\begin{table*}[!t]\centering
\caption{\textbf{Main results.} Closed-loop \NCAP{} decides; \emph{stat.\ / front.\ / side} are its per-scenario scores. Challenger is open-loop adversarial replay, and clean nuScenes checks that normal driving is preserved. \emph{ours} denotes our model-aware scenes. Stock rows evaluate the released checkpoints without fine-tuning; all other rows are means over \numSeeds{} seeds.}
\label{tab:main}
\small\setlength{\tabcolsep}{4.5pt}
\begin{tabular}{llccccccccc}\toprule
& & \multicolumn{5}{c}{\NCAP{} (closed-loop)} & \multicolumn{2}{c}{Challenger} & \multicolumn{2}{c}{nuScenes} \\
\cmidrule(lr){3-7}\cmidrule(lr){8-9}\cmidrule(lr){10-11}
Planner & Training data & score $\uparrow$ & CR (\%) $\downarrow$ & stat.\ $\uparrow$ & front.\ $\uparrow$ & side $\uparrow$ & CR (\%) $\downarrow$ & L2 $\downarrow$ & CR (\%) $\downarrow$ & L2 $\downarrow$ \\ \midrule
\multirow{5}{*}{\SD{}}
 & none (stock)        & \sdStockNcap{} & \sdStockCR & \sdStockStat{} & \sdStockFront{} & \sdStockSide{} & \sdStockChalCR{} & \sdStockChalL{} & \sdStockNuscCR{} & \textbf{\sdStockNuscL{}} \\
 & recovery            & \sdRecNcap{}\std{\sdRecNcapStd} & \sdRecCR\std{\sdRecCRStd} & \sdRecStat{} & \sdRecFront{} & \sdRecSide{} & \sdRecChalCR{}\std{\sdRecChalCRStd} & \sdRecChalL{}\std{\sdRecChalLStd} & \sdRecNuscCR{}\std{\sdRecNuscCRStd} & \sdRecNuscL{}\std{\sdRecNuscLStd} \\
 & rule-based          & \sdRuleNcap{}\std{\sdRuleNcapStd} & \sdRuleCR\std{\sdRuleCRStd} & \sdRuleStat{} & \sdRuleFront{} & \sdRuleSide{} & \sdRuleChalCR{}\std{\sdRuleChalCRStd} & \sdRuleChalL{}\std{\sdRuleChalLStd} & \sdRuleNuscCR{}\std{\sdRuleNuscCRStd} & \sdRuleNuscL{}\std{\sdRuleNuscLStd} \\
  & Challenger-style & \sdChalNcap{}\std{\sdChalNcapStd} & \sdChalCR{}\std{\sdChalCRStd} & \sdChalStat{} & \textbf{\sdChalFront{}} & \sdChalSide{} & \sdChalChalCR{}\std{\sdChalChalCRStd} & \textbf{\sdChalChalL{}}\std{\sdChalChalLStd} & \sdChalNuscCR{}\std{\sdChalNuscCRStd} & \sdChalNuscL{}\std{\sdChalNuscLStd} \\
 & \textbf{ours}       & \textbf{\sdOursNcap{}}\std{\sdOursNcapStd} & \textbf{\sdOursCR}\std{\sdOursCRStd} & \textbf{\sdOursStat{}} & \sdOursFront{} & \textbf{\sdOursSide{}} & \textbf{\sdOursChalCR{}}\std{\sdOursChalCRStd} & \sdOursChalL{}\std{\sdOursChalLStd} & \textbf{\sdOursNuscCR{}}\std{\sdOursNuscCRStd} & \sdOursNuscL{}\std{\sdOursNuscLStd} \\ \midrule
\multirow{5}{*}{\DD{}}
 & none (stock)        & \ddStockNcap{} & \ddStockCR & \ddStockStat{} & \ddStockFront{} & \ddStockSide{} & \ddStockChalCR{} & \ddStockChalL{} & \ddStockNuscCR{} & \textbf{\ddStockNuscL{}} \\
  & recovery & \ddRecNcap{}\std{\ddRecNcapStd} & \ddRecCR\std{\ddRecCRStd} & \ddRecStat{} & \ddRecFront{} & \ddRecSide{} & \ddRecChalCR{}\std{\ddRecChalCRStd} & \ddRecChalL{}\std{\ddRecChalLStd} & \ddRecNuscCR{}\std{\ddRecNuscCRStd} & \ddRecNuscL{}\std{\ddRecNuscLStd} \\
 & rule-based          & \ddRuleNcap{}\std{\ddRuleNcapStd} & \ddRuleCR\std{\ddRuleCRStd} & \ddRuleStat{} & \ddRuleFront{} & \ddRuleSide{} & \ddRuleChalCR{}\std{\ddRuleChalCRStd} & \ddRuleChalL{}\std{\ddRuleChalLStd} & \ddRuleNuscCR{}\std{\ddRuleNuscCRStd} & \ddRuleNuscL{}\std{\ddRuleNuscLStd} \\
  & Challenger-style & \ddChalNcap{}\std{\ddChalNcapStd} & \ddChalCR{}\std{\ddChalCRStd} & \ddChalStat{} & \ddChalFront{} & \ddChalSide{} & \textbf{\ddChalChalCR{}}\std{\ddChalChalCRStd} & \textbf{\ddChalChalL{}}\std{\ddChalChalLStd} & \ddChalNuscCR{}\std{\ddChalNuscCRStd} & \ddChalNuscL{}\std{\ddChalNuscLStd} \\
 & \textbf{ours}       & \textbf{\ddOursNcap{}}\std{\ddOursNcapStd} & \textbf{\ddOursCR}\std{\ddOursCRStd} & \textbf{\ddOursStat{}} & \textbf{\ddOursFront{}} & \textbf{\ddOursSide{}} & \ddOursChalCR{}\std{\ddOursChalCRStd} & \ddOursChalL{}\std{\ddOursChalLStd} & \textbf{\ddOursNuscCR{}}\std{\ddOursNuscCRStd} & \ddOursNuscL{}\std{\ddOursNuscLStd} \\
\bottomrule\end{tabular}
\end{table*}

When an E2E model collides with an adversarial agent, we ask whether the agent was undetected or detected and mishandled. We decompose every NeuroNCAP episode of the unmodified (stock) \DD{} by whether its own detections tracked the adversary before the collision. We count the adversary as detected when recall at 5--35\,m is at least 0.5.
Recall is computed on the detections the planner actually consumes, at the model's native operating point with no additional confidence threshold.
Table~\ref{tab:failure} shows the result.
Detection is reliable in every category, and in \failPlanning{} of the collision episodes the adversary was detected, while pure perception failures account for only 4.1\% of them. \SD{} decomposes the same way, with 95.8\% of its collisions occurring while the threat was detected.
The per-scenario pattern sharpens the diagnosis, as stationary threats are mostly handled properly, frontal interactions are crashed more often than avoided, and side scenarios collide in half of all episodes despite the best recall of the three.
The diagnosis does not hinge on the recall threshold, as sweeping it from 0.3 to 0.7 moves the detected-and-crashed share from 97.0\% to 92.5\% on \SD{} and 97.8\% to 89.3\% on \DD{}, while episodes that crash with the threat undetected never exceed 3.1\% of all episodes.
Detection also holds at close range, with recall rising from 0.84 beyond 15\,m to 0.91 within 15\,m on \DD{} and from 0.84 to 0.90 on \SD{}.
The bottleneck is therefore planning, which motivates freezing perception and fine-tuning only the planning layers.

\subsection{Main Results}\label{sec:exp-main}

\begin{figure*}[!t]\centering
\begin{minipage}[t]{0.492\textwidth}\centering
\includegraphics[width=\linewidth]{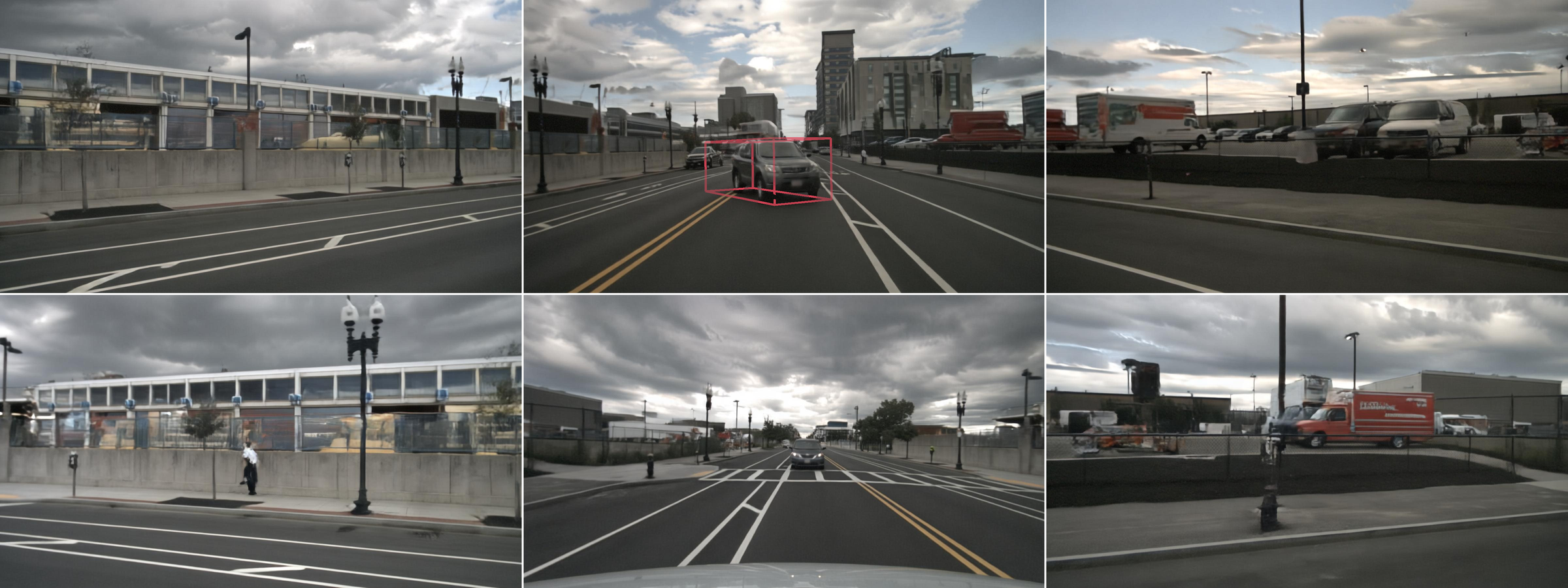}\\[5pt]
\includegraphics[width=\linewidth]{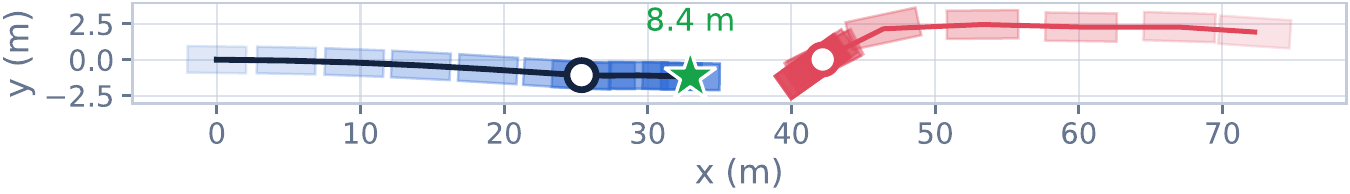}
\end{minipage}\hfill
\begin{minipage}[t]{0.492\textwidth}\centering
\includegraphics[width=\linewidth]{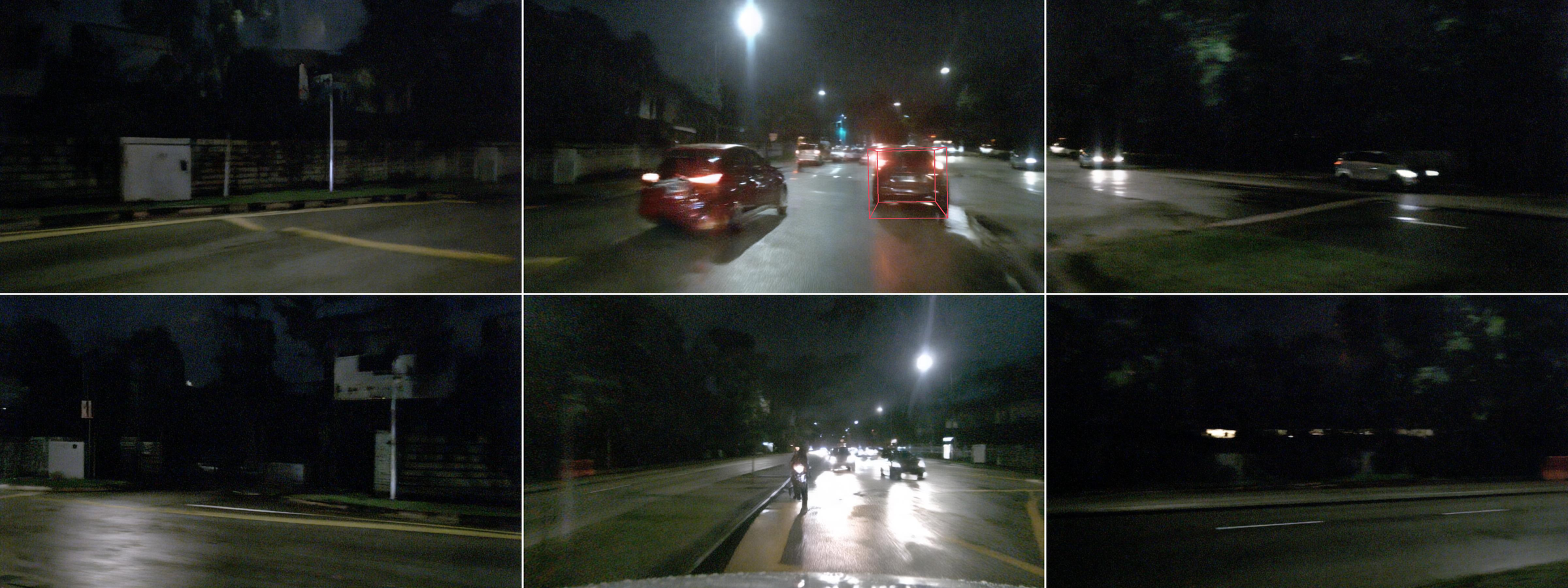}\\[5pt]
\includegraphics[width=\linewidth]{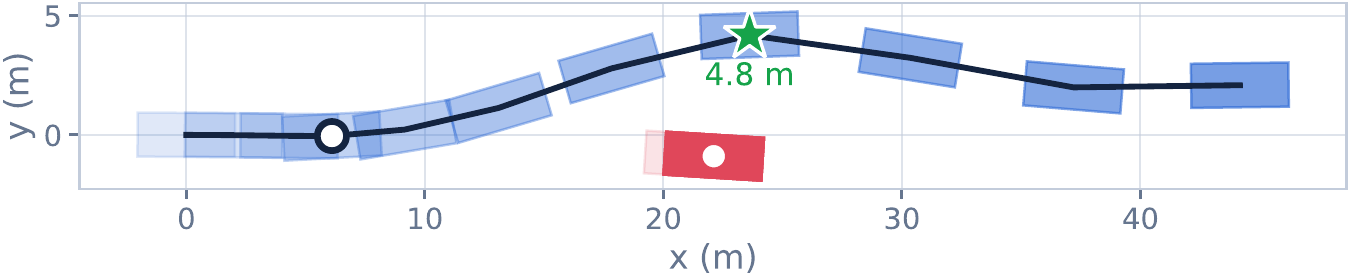}
\end{minipage}\\[2pt]
\includegraphics[width=0.99\textwidth]{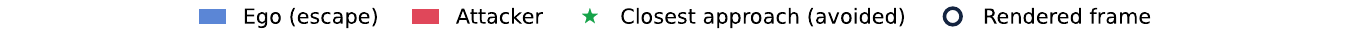}
\vspace{-5pt}

\caption{\textbf{Generated training scenes,} one per planner. Six cameras at the moment the ego must react, attacker boxed in red, with the BEV trail below, where footprints darken with time and the escape clears the attacker by the marked distance. Left: \DD{}-targeted in-path attack. Right: \SD{}-targeted front-brake at night, where the ego swerves around the near-stationary attacker.}
\label{fig:qualitative}
\end{figure*}

Fig.~\ref{fig:qualitative} shows two of the generated scenes, one per planner, a daytime in-path attack aimed at \DD{} and a night front-brake attack aimed at \SD{}. Each scene is rendered across all six cameras with the attack and its solved escape traced in BEV.
In addition, Table~\ref{tab:main} gives the quantitative results.
Our scenes roughly halve the closed-loop collision rate of both planners, \sdStockCR\pct{} to \sdOursCR\pct{} on \SD{} and \ddStockCR\pct{} to \ddOursCR\pct{} on \DD{}. Every individual seed improves the stock planner.
The per-scenario scores locate the gain.
Side collisions, the hardest category for both stock planners and the one our failure analysis flagged, improve by about 1.5 points on each, the largest gain of any category.
Frontal improves more modestly and remains the hardest category after fine-tuning. 
Stationary scores, already the highest category for both planners, increase further, and \SD{} reaches the maximum score of 5.0.
Normal driving is largely preserved. The nuScenes collision rate is unchanged on \SD{} (\sdOursNuscCR{} vs.\ \sdStockNuscCR{}) and slightly lower on \DD{} (\ddOursNuscCR{} vs.\ \ddStockNuscCR{}), and the nuScenes L2 is at most 0.06 higher than that of the stock planner (\sdOursNuscL{} vs.\ \sdStockNuscL{} for \SD{}, \ddOursNuscL{} vs.\ \ddStockNuscL{} for \DD{}).
On open-loop Challenger, our collision rate is lower than that of the stock planner on both planners, while all three baselines score better on \DD{}.
The larger L2 is expected, since the Challenger ground truth does not react to the adversary (Sec.~\ref{sec:exp-setup}) while our fine-tuned planners do.

Our model-aware data is also the best row in both planner blocks, ahead of all three baselines on both score and collision rate.
Recovery data lowers the collision rate from \sdStockCR\pct{} to \sdRecCR\pct{} on \SD{} and from \ddStockCR\pct{} to \ddRecCR\pct{} on \DD{}, a gain of 4.2 and 2.8 points against 16 to 18 points with our data. So photorealistic augmentation that varies the ego's own trajectory, without an adversarial vehicle, gives only a small gain.
Designing adversarial interactions helps, but neither scripted nor untargeted threats are sufficient.
Rule-based scenes correspond to two of \NCAP{}'s three categories, yet their collision rate is \sdRuleCR\pct{} on \SD{} and \ddRuleCR\pct{} on \DD{}, against \sdOursCR\pct{} and \ddOursCR\pct{} with our data, 15 points lower on \DD{}.
The difference is that our attacks are optimized against the planner's own head and cover the attack types at which that planner actually fails, rather than the types a script prescribes.
Challenger-style scenes, adversarial but aimed at no planner and labeled with the non-reacting logged trajectory, lower \SD{}'s collision rate to \sdChalCR\pct{} and leave \DD{} at \ddChalCR\pct{}, no better than stock. The same data gives the best Challenger L2 on both planners and the best Challenger collision rate on \DD{}. Open-loop Challenger metrics therefore do not track closed-loop safety on these planners.

\subsection{Choice of Escape Label}\label{sec:escape-abl}
\begin{table}[!t]\centering
\caption{\textbf{Choice of expert label.} Scenes, budget and recipe are identical; only the label changes. CEM-Diffusion refines samples from a nuPlan-trained trajectory diffusion prior~\cite{challenger,nuplan}, and brake-always applies maximum deceleration. \DD{}, mean over \numSeeds{} seeds.}
\label{tab:escape}
\small
\begin{tabular}{lcccc}\toprule
& \multicolumn{2}{c}{\NCAP{} (closed-loop)} & \multicolumn{2}{c}{\textsc{Challenger}} \\
\cmidrule(lr){2-3}\cmidrule(lr){4-5}
Expert label & score $\uparrow$ & CR (\%) $\downarrow$ & CR (\%) $\downarrow$ & L2 $\downarrow$ \\ \midrule
none (stock)           & \ddStockNcap{} & \ddStockCR & \ddStockChalCR{} & \ddStockChalL{} \\ \midrule
CEM-Diffusion          & \ddRuleNcap{}  & \ddRuleCR  & \textbf{\ddRuleChalCR{}} & \textbf{\ddRuleChalL{}} \\
brake-always           & \ddBrakeNcap{} & \ddBrakeCR & \ddBrakeChalCR{} & \ddBrakeChalL{} \\
\textbf{solved (ours)} & \textbf{\fsOursNcap{}} & \textbf{\fsOursCR} & \fsOursChalCR{} & \fsOursChalL{} \\
\bottomrule\end{tabular}
\end{table}

Table~\ref{tab:escape} compares three escape labels on the same rule-based scenes: the CEM-Diffusion expert, a brake-always label, and our solved escape.
The solved escape has the highest mean closed-loop score and the lowest collision rate, but its margin over brake-always is small.
Brake-always comes close in closed-loop because, with a single adversary and no cross traffic, stopping usually avoids the collision.
The three labels differ clearly on Challenger.
Brake-always has the highest Challenger collision rate, since the label must provide an evasive path rather than a stop.
CEM-Diffusion has the lowest Challenger collision rate and L2 but the lowest closed-loop score.
Even with our label, rule-based scenes remain below our scenes in Table~\ref{tab:main} (\fsOursNcap{} vs.\ \ddOursNcap{}); that difference is the contribution of the model-aware attacker.

\subsection{Is the Surrogate Faithful?}\label{sec:fidelity}
\begin{table}[!t]\centering
\caption{\textbf{Surrogate fidelity} on the \fidEligClean{} nuScenes val samples with annotated futures. Incidence is the fraction of samples with a collision within 3\,s. It is stricter than the nuScenes collision rate of Table~\ref{tab:main}, which averages over timesteps.}
\label{tab:fidelity}
\small
\begin{tabular}{lcc}\toprule
 & \SD{} & \DD{} \\ \midrule
$\planner$ vs.\ GT, L2 \emph{(the bar)} & \sdBarClean{} & \ddBarClean{} \\
$\pihat$ vs.\ $\planner$, L2            & \sdPihatClean{} & \ddPihatClean{} \\
\quad ratio to the bar                    & \sdPihatCleanX$\times$ & \ddPihatCleanX$\times$ \\ \midrule
collision \emph{incidence}, $\planner$  & \sdBarCR\pct & \ddBarCR\pct \\
collision \emph{incidence}, $\pihat$      & \sdPihatCR\pct & \ddPihatCR\pct \\\midrule
precision                                 & \sdPrecClean{} & \ddPrecClean{} \\
\quad lift over base rate                 & \textbf{\sdLift} & \textbf{\ddLift} \\
recall                                    & \sdRecallClean{} & \ddRecallClean{} \\
\bottomrule\end{tabular}
\end{table}

The surrogate does not need to reproduce the planner in every situation.
Its purpose is to find the scenes in which the planner fails.
We therefore evaluate two properties.
The first is how far $\pihat$ deviates from the planner, relative to how far the planner deviates from ground truth.
The second is how reliably a collision predicted by $\pihat$ corresponds to a collision of the real planner.
Table~\ref{tab:fidelity} reports both on nuScenes val, using real images only, because on rendered scenes the comparison would mix adapter error with any drift of the planner's features $c$ on synthetic pixels.

On the first property, $\pihat$ deviates from its planner by about as much as the planner deviates from ground truth, \sdPihatCleanX{} times that deviation for \SD{} and \ddPihatCleanX{} times for \DD{}.
On the second, we compare collision \emph{incidence}, which counts a sample as colliding if any predicted timestep collides.
Incidence is a per-sample quantity and differs from the per-timestep collision rate of Table~\ref{tab:main}.
A collision predicted by $\pihat$ corresponds to a real collision with precision \sdPrecClean{} on \SD{} and \ddPrecClean{} on \DD{}, against base rates near 1.5\%, i.e., 38 and 51 times the base rate.
Recall is \sdRecallClean{} on \SD{} and \ddRecallClean{} on \DD{}.
For a scene generator, low recall is acceptable, since a missed failure mode only reduces coverage.
On the rendered attack scenes themselves, the stock planner, which has never seen them, collides with the attacker in 32 of 50 \SD{} scenes and 26 of 50 \DD{} scenes. Scenes the stock planner survives still supply a corrective label, since the solved escape maximizes clearance.
The transfer experiment in Sec.~\ref{sec:premium} provides further evidence of fidelity.

\subsection{The Targeting Premium Is Planner-Dependent}\label{sec:premium}
\begin{table}[!t]\centering
\caption{\textbf{Transfer and targeting premium.} Each planner fine-tuned on data generated against itself (diagonal) and against the other (off-diagonal). Cells give closed-loop \NCAP{} score / collision rate, mean over \numSeeds{} seeds.}
\label{tab:transfer}
\small
\begin{tabular}{lcc}\toprule
data $\downarrow$ \ / \ model $\rightarrow$ & \SD{} & \DD{} \\ \midrule
\SD-targeted & \sdOursNcap{} / \sdOursCR\pct & \ddCrossNcap{} / \ddCrossCR\pct \\
\DD-targeted & \sdCrossNcap{} / \sdCrossCR\pct & \ddOursNcap{} / \ddOursCR\pct \\ \midrule
premium (own $-$ cross) & $+\premiumSD$ & $+\premiumDD$ \\ \bottomrule
\end{tabular}\end{table}

If the surrogate's gradients carry planner-specific information, data generated against planner $A$ should help $A$ more than it helps $B$.
Table~\ref{tab:transfer} shows the full two-by-two comparison, in which each planner is fine-tuned on its own scenes and on the scenes of the other planner, with \numSeeds{} seeds per entry.
We call the difference between the own-data score and the cross-data score the \emph{targeting premium}.
Both premiums are positive.
Every own-data seed scores higher than every cross-data seed on both planners.
The premiums nonetheless differ, $+\premiumSD$ for \SD{} against $+\premiumDD$ for \DD{}.
In terms of collision rate, \DD{} trained on \SD{}'s scenes keeps only 6.7 of the 18.3-point reduction it obtains on its own scenes.

\begin{figure}[!t]
  \centering
  \includegraphics[width=\columnwidth]{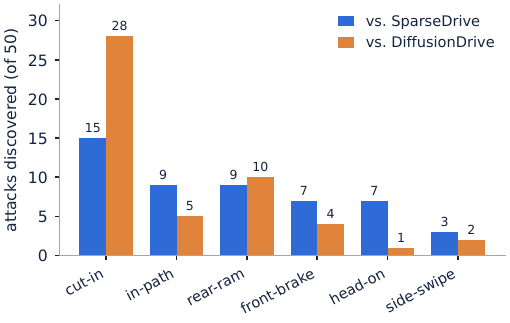}
  \vspace{-10pt}
    \caption{\textbf{Attacks discovered against each planner differ.} Those against \DD{} are predominantly cut-ins and almost never head-on threats, while those against \SD{} also include head-on and front-brake attacks.}
  \label{fig:mechanism}
\end{figure}

Fig.~\ref{fig:mechanism} shows the distribution of attack types found against each planner.
The attacks against \DD{} concentrate on cut-ins with almost no head-on threats, while those against \SD{} are spread over cut-ins, head-on and front-brake.
Since every attack is optimized against one planner's head, attacks of the same type are also expected to differ between the two planners.
The transfer result reflects this, since the scenes found through each surrogate help that surrogate's own planner most, which requires that the gradients behind those scenes encode failure modes specific to that planner.
A planner-independent proxy would not produce this asymmetry.

\subsection{Emergent Diversity}\label{sec:diversity}
The pipeline does not specify scenario types, yet six attack types emerge from the single collision objective (Fig.~\ref{fig:mechanism}).
We assign each attack to a type with a fixed geometric rule based on the adversary's position and heading relative to the ego at the moment of first contact and during the approach.
The proportions of the types differ between the two planners.
\DD{}'s concentration on cut-ins reflects a weakness of that planner rather than a lack of diversity in the search, since the same search against \SD{} produces a wider spread.
The escapes are similarly varied and include braking, holding speed, swerving, and accelerating past the adversary, with several qualitatively distinct valid escapes per scene on average.
The main results use only the escape with the largest clearance; the remaining escapes are left for future multi-label training.
\section{Limitations}\label{sec:limitations}
The first limitation is breadth.
Our evidence covers two planners, so the targeting premium is measured at two points only, and a third architecture would be needed to confirm the trend.
The renderer also restricts us to nuScenes-based planners, and we evaluate on a single closed-loop simulator.
\NCAP{}'s neural rendering differs from the generative renderer behind our training data, so the gains are not specific to one renderer, but evaluation on further simulators would strengthen this claim.
The second limitation is depth. \DD{}'s surrogate has a recall of \ddRecallClean{} on that planner's real collisions (Table~\ref{tab:fidelity}), so it misses roughly half of them, and the attacks we find are a subset of the attacks that would succeed against the real planner. The measured premium is therefore a lower bound on what a better surrogate would expose.
The third limitation is the scope of the scenes.
Each scene contains a single adversarial vehicle and no vulnerable road users. We plan to extend the search to multiple adversaries and to vulnerable road users in the future.

Two problems remain open. Frontal scenarios remain the hardest category after fine-tuning. We also train on one escape per scene although the solver produces several, and training with multiple labels per scene is left for future work.
\section{Conclusion}
This paper began with a failure analysis of current E2E models.
Decomposing closed-loop collisions showed that end-to-end planners detect the threat in the large majority of collisions but fail to respond to it. The failures are thus mainly a planning problem and not a perception problem, and the missing safety-critical training data must target planning.
The obstacle to generating such data against a specific planner is that a vision-based planner consumes images, and a video renderer inside an optimization loop is computationally infeasible.
Distilling the planner into a differentiable surrogate built from its own frozen planning head, but operating on BEV inputs instead of images, removes this obstacle.
With the renderer outside the optimization loop, gradients are cheap and the attacks are specific to one planner.

Fifty such scenes per planner, each pairing an optimized attack with a safe escape label, roughly halve the closed-loop collision rate of two vision-based planners while normal driving is largely preserved, at around 30 GPU-hours of search per planner.
The transfer experiment shows that this gain depends on the planner the scenes were generated for.
Each planner improves most on the scenes generated against itself, on every seed, and \DD{} keeps only about a third of its collision-rate reduction when trained on scenes generated against \SD{}.
This asymmetry is consistent with the method.
Safety-critical training data is therefore planner-specific, and generating it against the planner's own surrogate is an efficient way to obtain it.



\bibliographystyle{ieeetr}
\bibliography{refs}
\end{document}